\documentclass{article}
\usepackage{ijcai25}

\usepackage{times}
\usepackage{soul}
\usepackage{url}
\usepackage{color}
\usepackage[hidelinks]{hyperref}
\usepackage[utf8]{inputenc}
\usepackage[small]{caption}
\usepackage{graphicx}
\usepackage{amsmath}
\usepackage{amsthm}
\usepackage{booktabs}
\usepackage{algorithm}
\usepackage{algorithmic}
\usepackage[switch]{lineno}
\usepackage{subfigure}

\title{Unsupervised Feature Transformation via In-context Generation, Generator-critic LLM Agents, and Duet-play Teaming}

\author{
Nanxu Gong$^1$
\and
Xinyuan Wang$^1$\and
Wangyang Ying$^1$\and
Haoyue Bai$^1$\and
Sixun Dong$^1$\and
Haifeng Chen$^2$\and
Yanjie Fu$^1$\\
\affiliations
$^1$School of Computing and Augmented Intelligence, Arizona State University\\
$^2$NEC Laboratories America\\
\emails
\{nanxugong, xwang735, wying4, haoyueba, sixundong, yanjie.fu\}@asu.edu,
haifeng@nec-labs.com
}

\begin{document}

\maketitle
\begin{abstract}
Feature transformation involves generating a new set of features from the original dataset to enhance the data's utility.
In certain domains like material performance screening, dimensionality is large and collecting labels is expensive and lengthy. 
It highly necessitates transforming feature spaces efficiently and without supervision to enhance data readiness and AI utility.
However, existing methods fall short in efficient navigation of a vast space of feature combinations, and are mostly designed for supervised settings. 
To fill this gap, our unique perspective is to leverage a generator-critic duet-play teaming framework using LLM agents and in-context learning to derive pseudo-supervision from unsupervised data.
The framework consists of three interconnected steps: (1) Critic agent diagnoses data to generate actionable advice, (2) Generator agent produces tokenized feature transformations guided by the critic’s advice, and (3) Iterative refinement ensures continuous improvement through feedback between agents.
The generator-critic framework can be generalized to human-agent collaborative generation, by replacing the critic agent with human experts. 
Extensive experiments demonstrate that the proposed framework outperforms even supervised baselines in feature transformation efficiency, robustness, and practical applicability across diverse datasets. Our code is publicly available at \href{https://github.com/NanxuGong/LPFG}{https://github.com/NanxuGong/LPFG}.
\end{abstract}
\vspace{-0.3cm} 
\section{Introduction} 
Feature transformation aims to rebuild a new feature space from an original feature set (e.g. $[f_1, f_2] \rightarrow [\frac{f_1}{f_2}, f_1-f_2, \frac{f_1+f_2}{f_1}]$). 
Feature transformation can advance the  power (structural, predictive, interaction, and expression levels) of data to make data AI ready.
%Traditional feature transformation is often inefficient due to large search space, and require task-specific supervised labels to evaluate and guide transformation. 
In many practices, feature transformation is conducted either by human experts, or by machine-assisted search guided through downstream task feedback.
In certain domains, like material synthesis and performance screening, feature transformation is particularly useful in capturing interactions and compositions within material formulas to identify performance drivers. 
However, 1) there are millions of candidate material ingredients (i.e., features) for material synthesis, thus, it is inefficient to explore and search all feature combinations and interaction possibilities; 2) obtaining supervised material performance labels often requires time-intensive and costly in-lab experiments.
This practical challenge highlights the need for a new AI task: efficient and unsupervised feature transformation (EUFT). 

There are two major challenges in solving EUFT: 1) efficient transformation, and 2) unsupervised transformation. 
Firstly, there is an exponentially expanding range of feature combination possibilities in a feature space, leading to an overwhelmingly large discrete search space. Efficient transformation is to answer: how can we avoid searching a large feature combination space when generating feature transformations?
Secondly, in supervised settings, most methods exploit predictive accuracy feedback of a transformed feature set on a downstream ML model to guide optimal feature transformation search. 
Under unsupervised settings, there is no supervised knowledge as guidance.
Unsupervised transformation aims to answer: how can we discover supervision knowledge from unsupervised data to steer the optimal feature transformation generation?

There are significant gaps in current methodologies for EUFT.
1) \textit{Manual feature transformation} requires domain and empirical expertise to formulate task-specific strategies, thus is inefficient and doesn't generalize well in unsupervised settings. 
2) There are studies that solve feature transformation as \textit{discrete or continuous search tasks}~\cite{wang2022GRFG,wang2023reinforcement}. 
Technical solutions include reinforcement learning, genetic algorithm, and generative learning based reformulations. They either search optimal feature sets in a discrete or continuous space. However, the models are time-consuming and require downstream supervised feedback to guide search.  
3) \textit{LLM agent based methods}~\cite{gong2024evolutionary,zhang2024dynamic,hollmann2024large}  interpret the prompt using their pretrained general and world token knowledge to generate outputs that align with the given patterns, scores, or comparisons patterns of features to generate tokenized feature transformations. But, existing methods target at supervised settings, instead of unsupervised setting; they regard LLM as a generator and ignore its other abilities: in-context learning enabled teaming, diagnosis and critics, duet-play can deliver better feature transformations in more challenging computational or learning settings. 

\textbf{Our insights: a duet-play generator-critic teaming perspective to derive supervision from unsupervised data.} 
We highlight two research insights. 
To address  efficient transformation, we show that we can tokenize a set of transformed features into a feature cross token sequence, thereafter see LLM as a generator to learn patterns from the text they process, generate feature transformation tokens, and avoid searching in a large combination space. 
To address unsupervised transformation,  we found that LLM agents exhibit a feature space diagnosis ability over tokenized data. We use such diagnosis ability to derive supervision knowledge from unsupervised data to guide the feature transformation process. 
We propose to unify generator-critic agents, in-context learning,  and duet-play teaming to create ``pseudo model'', ``pseudo objective'', and ``pseudo optimization'' with only unlabeled data. 
In particular, the pseudo model is the generator agent that generates feature transformations given a dataset. 
The pseudo objective is the critic agent that diagnoses a dataset to generate feature space improvement advices as ``textual gradient'', which is equivalent to deriving optimization direction (i.e., gradient descent) from unsupervised data. 
The pseudo optimization is duet-play teaming between the critic agent and the generator agent, in which the advices of the critic agent are utilized to augment the in-context learning prompt of the generator agent in order to transform better feature space. The two agents team together and iteratively duet-play the same process. 

\textbf{Summary of proposed approach.} 
We propose a duet-play generator-critic agent teaming framework to derive supervision from unsupervised data for fast unsupervised feature transformation.
The framework includes three steps: 
1) \textit{The critic step:} the critic agent diagnoses semantic relationships and data distribution properties to generate advice for improving feature spaces; 
2) \textit{The generation step:} the generator agent tokenizes features, operators, and transformations and leverages in-context learning to produce a tokenized transformed feature set based on critic agent augmented prompts; 
3) \textit{Iterative refinement:} A feedback loop between the critic agent and generator agents ensures continuous improvement of the generated features through semantic and structural alignment. 
In addition, this framework can be generalized from critic-augmented generation to human-agent collaborative generation, by replacing the critic agent with human experts. 
Finally, extensive experiments demonstrate our method is extremely efficient while accurate, highlighting its practical potential for EUFT.

\section{Problem Definition}
% Formally, in the feature generation task, we define a dataset $\mathcal{D} = \{X,y\}$, where $X = [f_1,,f_2, ..., f_n]$ is the original feature set and $y$ is the label. There is also an operator set $\mathcal{O}$ including mathematical operators such as $log, cube, +, *$. In our framework, we firstly apply an optimizer LLM $f$ to evaluate the dataset and give some advice on improving the feature space, expressed by: $\theta = f (X)$. Based on the prior knowledge on data science and advice from the optimizer, we further conduct feature generation by employing a generator LLM with the parameter $g$. The generated feature set is defined by: $\mathcal{\hat{X}} \sim g(\cdot|X,\theta)$.
We utilize LLMs as agents for unsupervised feature generation. This approach enables the automated generation of meaningful features and improves performance in downstream tasks. Formally, given a dataset $\mathcal{D} = \{X,y\}$ and an operation set $\mathcal{O}$, where $X$ is the original feature set and $y$ is the corresponding label. Here, $y$ is only used for the testing process. Our framework employs a two-stage approach for feature generation using LLMs. First, an critic LLM analyzes the original data set and then provides generation feedback to guide how to enhance the feature space, represented as $\theta = k(X)$, where $k$ is the function notation of the critic. Second, we employ a generator LLM to perform feature generation. The generating feature space is represented by $\mathcal{\hat{X}} \sim g(\cdot|X, \mathcal{O}, \theta)$, which incorporates the original features and the critic insights. This two-stage process is repeated iteratively, with each step refining the feature space to enhance its representation for downstream tasks.

\section{Generator-critic Feature Transformation}

\begin{figure*}[t]
    \centering
    \includegraphics[width=\linewidth]{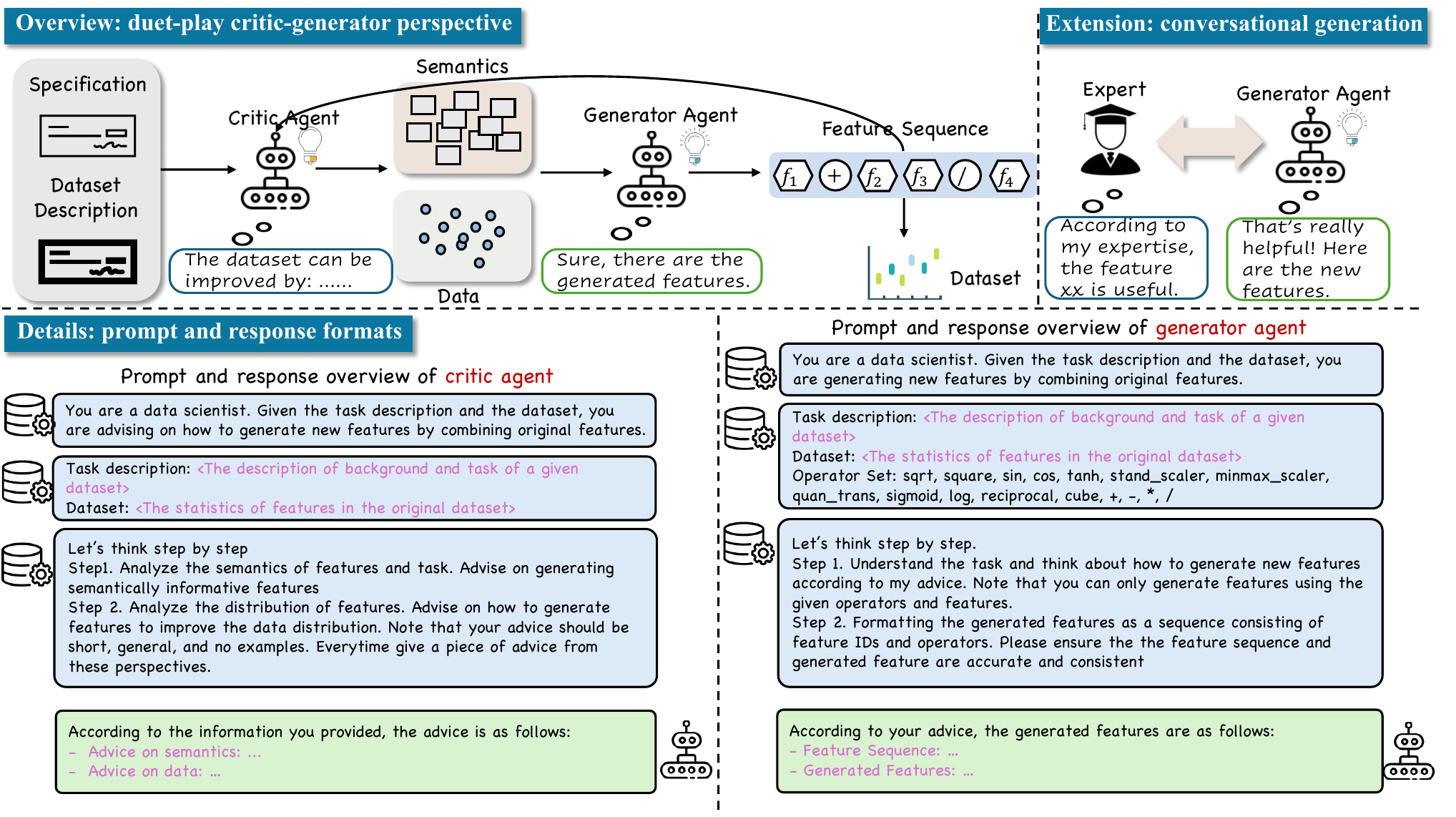}
    \caption{Framework overview. We implement feature generation through a duet-play generator-critic framework. We also extend it to a conversational generation manner.}
    \label{fig:framework}
    \vspace{-0.4cm}
\end{figure*}

\subsection{Framework Overview}
Figure~\ref{fig:framework} shows our generator-critic LLM agents framework for unsupervised duet-play feature transformation includes:
1) the critic step: develop diagnosis and advice on generating meaningful features; 
2) the generation step: feedback-driven feature generation; 
3) the iterative refinement step.

In \textbf{Step 1}, given a data set, the critic agent aims to diagnose the dataset from both semantic and structural perspectives. 
Our idea is to leverage the general knowledge and reasoning abilities of LLMs to uncover insights into feature interactions, data relationships, and potential strategies for data augmentation. Specifically, the critic agent  performs two key analyses: (i) \textit{Semantic Analysis}: examining feature descriptions and task objectives to derive meaningful interactions and transformations. (ii) \textit{Structural Analysis}: assessing the distribution and completeness of features to reason  how to transform a feature space that aligns with the downstream task. Step 1 is to output a textual description of how to transform a dataset to make it AI ready. The benefits of this step is that the semantic, structure, distribution-aware feature space transformation advices can inspire the generator agent into precise directions for generating informative and relevant features.

In \textbf{Step 2}, the generator agent regards a feature as a token,  an operator as a token, and a transformed feature as a token segment (e.g., $f_1*f_2$). Building upon this symbolic representation, the generator agent tokenizes a set of transformed features as a token sequence $(f1*f2),log(f3),(f4/f5)$. The task of feature transformation is reformulated into generating a feature transformation token sequence given a dataset, which is achieved by the generator agent.
% These sequences are represented in a compact and traceable postfix format, in order to
This formulation enables straightforward reconstruction of a transformed dataset from the original dataset. 
The benefits of designing the generator agent are (i) efficiency: in-context prompting ensures rapid generation without sacrificing quality; (ii) adaptability: the generator agent can dynamically adapt to diverse datasets and leverage both prior knowledge and task-specific feedback from the critic agent; (iii) traceability: the idea of using symbolic token sequences to represent actionable feature transformations and using GenAI to learn and generate can facilitate the transparency and reproducibility of feature transformations. 

In \textbf{Step 3}, to improve the robustness of the generated features, our framework iterates a feedback loop between the critic LLM agent and generator LLM agent. By comparing generated features with semantic rules and structural patterns identified by the critic LLM agent, we iteratively refine the feature space by coordinating the critic agent and the generator agent to converge toward ensuring semantic coherence, structural integrity,  predictive utility, and format compatibility with downstream tasks, all in an unsupervised setting.

\subsection{The Critic Step}

\textbf{Why the critic agent matters?} 
Firstly, a feature space is complex because features  can vary in dimensionality (e.g., high dimensionality), correlation (e.g., nonlinear, interdependent, or context-specific), type (e.g., categorical, numerical, ordinal, ratio), and information properties (e.g., scale, redundancy, noise, or overfitting). This complexity makes it challenging to capture meaningful patterns, thus, it highly necessitates an automated tool to optimize meaningful feature representations.
Secondly, traditional feature transformation relies on supervised feedback (e.g.,  task-specific predictive accuracy or feature importances) to guide optimization. 
In unsupervised settings, there are no explicit labels, making supervised feedback and optimization direction unavailable. 
The critic agent bridges this gap by evaluating datasets comprehensively, offering interpretable advice from both semantic and data distribution perspectives.

\noindent\textbf{Leveraging AGI diagnosis of feature space as ``textual gradient'' of in-context feature transformation optimization.} 
One strategy is to use a single LLM agent for generative feature transformation. However, the single agent needs to implicitly achieve both reasoning (i.e., diagnosing issues of a feature space and identifying improvement directions of feature space) and  generation (i.e., generating token sequences of symbolic feature transformation actions). 
This strategy introduces uncertainty in coupling reasoning and generation toward optimal and requires the  agent to precisely self-identify optimization direction.   
Our idea is to leverage a critic LLM agent to generate issue diagnosis and improvement advice of feature space as textual gradients, in order to provide unique optimization direction contexts within the prompt to adjust the generation agent's behavior. 

\noindent\textbf{Step 1: Semantic Diagnosis of Feature Space.} 
The critic agent, trained on extensive textual corpora, can infer semantic relationships between input variables and target outputs. In particular,  given the true names of predictors (X) and the response (Y), we prompt the critic agent to capture the semantic connections and contexts between X and Y to inspire the generator agent to create effective and interpretable features. 
% \textbf{\textcolor{red}{give an example of semantic diagnosis generated by the critic agent}}

\noindent\textbf{Step 2: Distribution Diagnosis of Feature Space.} 
Besides, we prompt the critic agent to evaluate the underlying data distributions of feature space, in order to perceive whether the classification patterns or decision boundaries of data are discriminative and easy to learn. 
We exploit the perceived distribution information to inspire the critic agent to think how features can be transformed to reshape data distributions so classification patterns are well separated. 
% \textbf{\textcolor{red}{give an example of distribution diagnosis generated by the critic agent}}

\begin{figure}[htb]
    \centering
    \includegraphics[width=\linewidth]{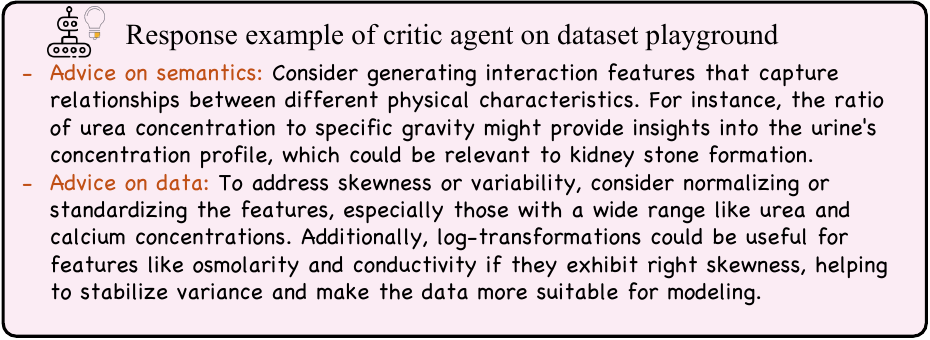}
    \caption{We provide a response example of critic agent on the dataset playground. We obtain advice from semantic and data perspectives. }
    \label{fig:res_cr}
    \vspace{-0.4cm}
\end{figure}

\noindent\textbf{Integrating Semantic and Distributional Diagnosis.} 
By integrating both the semantic diagnosis and the distributional diagnosis, the critic agent delivers well-rounded and actionable contexts to augment the prompt of the generator agent for feature space improvement. 
% \textbf{\textcolor{red}{''Let’s think step by step. Step 1. Analyze the semantics of features and task. Advise on generating semantically informative features. Step 2. Analyze the distribution of features. Advise on how to generate features to improve the data distribution. Note that your advice should be short, general, and no examples. Everytime give a piece of advice from these perspectives.  Format for Response:
% - Advice on semantics: …
% - Advice on data: …
% ''
% }}
In this way, we can ensure that feature generation aligns with both the structural properties and semantic relationships of the dataset. Figure \ref{fig:res_cr} presents a response example of the critic agent.

\subsection{The Generation Step}

\textbf{Why a generator agent matters?} 
In prior literature, feature transformation achieved by manual transformations, supervised transformations (e.g., reinforcement, evolutionary), unsupervised transformations (e.g., PCA). However,  manual methods are not generalizable and incomplete, as they heavily rely on domain and empirical experiences. Supervised methods require labeled data and need to search a vast space. Unsupervised methods are based on a strong assumption of straight linear feature correlation.
The success of LLM shows it is appealing to model language knowledge as token sequences and  reformulate predictive tasks as generative AI to regress next token. 
Following a similar spirit, we propose to represent mechanism-unknown feature space knowledge into symbolic sequential tokens. For instance, a transformed feature set ({$\frac{f_1}{f_2}, f_1-f_2, \frac{f_1+f_2}{f_1}$}) is seen as a feature operator cross token sequence ``$(f_1/f_2), (f_1-f_2), ((f_1+f_2)/{f_1})$EOS''. In other words, feature transformation can be viewed as a token generation task. Moreover, LLM exhibits in-context and few-shot learning abilities, thus, we can teach LLM to learn feature knowledge by demonstrating a list of feature transformation sequence examples in an instructional prompt. 

\noindent\textbf{Integrating feature-operator cross tokenization and in-context learning for fast and unsupervised feature transformation.} 
Our idea is to see features and operators as tokens , and a transformed feature as a token segmentation of feature-operator crosses. We regard feature transformation as a generative AI task. LLM is specialized in sequential token generation. Its in-context learning ability allows us to incorporate critic LLM to learn complex feature space knowledge and the optimization direction of sequence generation, by demonstrating relevant background information, specific instructions, examples of desired outputs, clear task definition, and structured formatting. Such reformulation can help to achieve efficient and unsupervised feature transformation, 

\noindent\textbf{Step 1: Tokenize Feature Transformations.} 
The generator agent tokenizes each transformation into a sequence of tokens (e.g., $f_1*f_2/f_3$). This symbolic representation not only enables the LLM token generation of feature transformation, but also facilitates the transparent tracking of transformations.

\begin{figure}[htb]
    \centering
    \includegraphics[width=\linewidth]{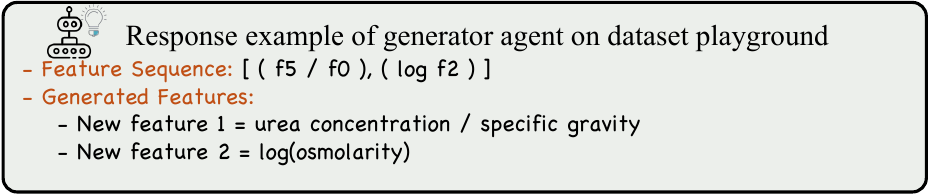}
    \caption{We present a response example of generator agent on the dataset playground. The feature sequence represent a dataset and the generated features interpret the semantic meanings.}
    \label{fig:res_ge}
    \vspace{-0.4cm}
\end{figure}

\noindent\textbf{Step 2: In-context Prompting for Rapid Generation.} 
The generator agent leverage the outputs from the critic agent to construct in-context prompts to dynamically generate high-quality features. By combining the task-specific guidance from the critic agent with the generic knowledge knowledge of the generator agent, the generator agent generates feature transformations without supervised labels. Figure \ref{fig:res_ge} demonstrates an example of the generator agent response.
% \textbf{\textcolor{red}{give an example of generation prompt}}

There are two major benefits of using the generator agent: 1) \textbf{Efficiency:} in-context learning ensures rapid feature generation without compromising on quality.  2) \textbf{Traceability:} The use of symbolic token sequences enhances transparency and reproducibility, making the feature transformations easily interpretable and verifiable.

\subsection{The Iterative Refinement Step}
\noindent\textbf{Leveraging critic agent-augmented generation for iterative improvement.} In each iteration, the critic agent generates semantic and distributional diagnosis of feature space, along with feature transformation advices, as enriched contexts. We then leverage the critic agent-generated diagnosis and advices to augment the in-context learning prompt of the generator agent.  During such iterative refinement,  the generator agent dynamically adapts to diverse datasets by integrating task-specific diagnosis and advices from the critic agent and AGI knowledge.

% \begin{figure}[h]
%     \centering
%     \includegraphics[width=\linewidth]{img/prompt.pdf}
%     \caption{Prompt outline. Optimizer give advice from the perspectives of semantics and data. Generatro creates new features according to the advice.}
%     \label{fig:prompt}
%     \vspace{-0.4cm}
% \end{figure}

\subsection{From Critic-Augmented Generation to Human-Agent Collaborative Generation}
The emergence of Reinforcement Learning from Human Feedback (RLHF) demonstrates the significance of human feedback for diverse and insightful generation \cite{yu2024rlhf,wang2023rlhf}. 
Traditionally, feature transformation is effective with domain knowledge and empirical experiences under the guidance of human experts.

We want to highlight that our critic-generator framework  can be converted into human-agent collaborative generation. Specifically,  we can replace the critic agent with a user or a human expert to implement customized feature transformation. The human expert can write domain-specific instructions to help LLM to incorporate human thinking and expert knowledge into the in-context learning. For instance, human expert can provide a domain specific and personalized instruction to LLM, by inputting an instruction like: "Feature $f_3$ is interesting. Please generate new variants of $f_3$." 
\begin{table*}[t]
\centering
\resizebox{0.8\linewidth}{!}{
\begin{tabular}{c|ccccccc|c}
\toprule
Dataset & Source & Original & TTG & AutoFeat & GRFG & OpenFE & CAAFE & LPFG \\ \midrule
balance & OpenML & 0.750 & 0.781 & 0.688 & \underline{0.813} & 0.781 & 0.781 & \textbf{0.906} \\ 
cmc & OpenML & 0.507 & 0.518 & 0.504 & \underline{0.542} & 0.534 & 0.507 & \textbf{0.561} \\ 
credit-g & OpenML & 0.780 & 0.78 & 0.772 & \underline{0.788} & 0.764 & 0.772 & \textbf{0.792} \\ 
diabetes & OpenML & 0.786 & \underline{0.797} & 0.786 & \underline{0.797} & \underline{0.797} & 0.786 & \textbf{0.813} \\ 
tic-tac-toe & OpenML & 0.667 & 0.708 & 0.625 & 0.750 & 0.625 & \underline{0.792} & \textbf{0.833} \\ 
pc1 & OpenML & \underline{0.935} & \textbf{0.939} & \underline{0.935} & \textbf{0.939} & \underline{0.935} & \textbf{0.939} & \textbf{0.939} \\ 
airlines & OpenML & 0.638 & 0.640 & 0.628 & \underline{0.644} & 0.614 & 0.624 & \textbf{0.650} \\ 
jungle & OpenML & 0.848 & 0.852 & \underline{0.856} & \underline{0.856} & 0.850 & 0.846 & \textbf{0.860} \\ \midrule
health & Kaggle & 0.742 & 0.740 & 0.740 & 0.742 & \underline{0.748} & 0.736 & \textbf{0.752} \\ 
pharyngitis & Kaggle & 0.680 & 0.711 & 0.680 & \textbf{0.719} & 0.695 & \underline{0.711} & \textbf{0.719} \\ 
spaceship & Kaggle & 0.744 & 0.754 & 0.746 & 0.748 & \textbf{0.756} & \underline{0.754} & 0.750 \\ 
playground & Kaggle & \underline{0.750} & \underline{0.750} & 0.721 & \underline{0.750} & 0.712 & 0.712 & \textbf{0.760} \\ \bottomrule
\end{tabular}}
\caption{Overall comparison of different models on across 12 datasets. We bold the best results and underline the second-best results.}
\label{tab:overall}
\end{table*}

\section{Experimental Results}

\subsection{Experimental Setup}
\textbf{Data Descriptions.} We utilize 12 public datasets that contain task descriptions and feature names from Kaggle and OpenML to conduct experiments. 

\noindent\textbf{Evluation Metrics.} We employ Random Forest as the downstream ML model. The accuracy score of the predictions is used to evaluate the performance of methods.

\noindent\textbf{Baselines and variants.} To demonstrate the effectiveness of our method, We compare LPFG with 5 widely-used models in feature transformation: 1) \textbf{TTG} \cite{khurana2018feature} formulates feature transformation as a graph and searches via reinforcement learning; 2) \textbf{AutoFeat} \cite{horn2020autofeat} expands the feature space and performs feature selection to retain the meaningful features; 3) \textbf{GRFG} \cite{wang2022GRFG} builds a multi-agent framework to automatically generate new features and optimize; 4) \textbf{OpenFE} \cite{zhang2023openfe} proposes a feature boosting method and a two-stage pruning algorithm to implement expand-reduce feature engineering; 5) \textbf{CAAFE} \cite{hollmann2024large} leverages the in-context learning ability of LLM to conduct feature transformation.

To comprehensively evaluate the necessity of each component of LPFG, we introduce variant models: 1) \textbf{LPFG-a} adopt the supervised performance of a given feature set on the downstream ML model as the feedback; 2) \textbf{LPFG-i} leverage the importance obtained from Random Forest to guide feature transformation; 3) \textbf{LPFG-o} remove the critic agent and assign the generator to handle both reasoning and generation.

\subsection{Experimental Results}
\textbf{Overall Performance.}  This experiment aims to answer: \textit{Is the proposed method effective for improving downstream ML model performance?} We compare the proposed method with the baselines on 12 datasets. Note that the LLM-based methods (i.e., CAAFE and LPFG) use GPT-3.5-turbo through API. Table \ref{tab:overall} shows LPFG, as an unsupervised method, outperforms all supervised baselines on most datasets. The underlying driver is that the critic agent can comprehensively evaluate the datasets and give useful advice for feature transformation. Besides, an intriguing observation from the dataset \textit{playground} reveals that LPFG is capable of identifying effective optimization directions even on challenging datasets, whereas the baseline methods fail to improve performance. On the one hand, that is probably because the generator-critic framework is more noise-resistant and robust. On the other hand, a potential distribution shift may affect the performance of supervised methods. However, LPFG demonstrates the capability to achieve more reliable optimization by conducting a comprehensive evaluation of the dataset.

\begin{figure}[h]
    \centering
    \subfigure[LLM Impact]{
    \begin{minipage}[ht]{0.472\linewidth}
    \centering
    \includegraphics[width=1.6in]{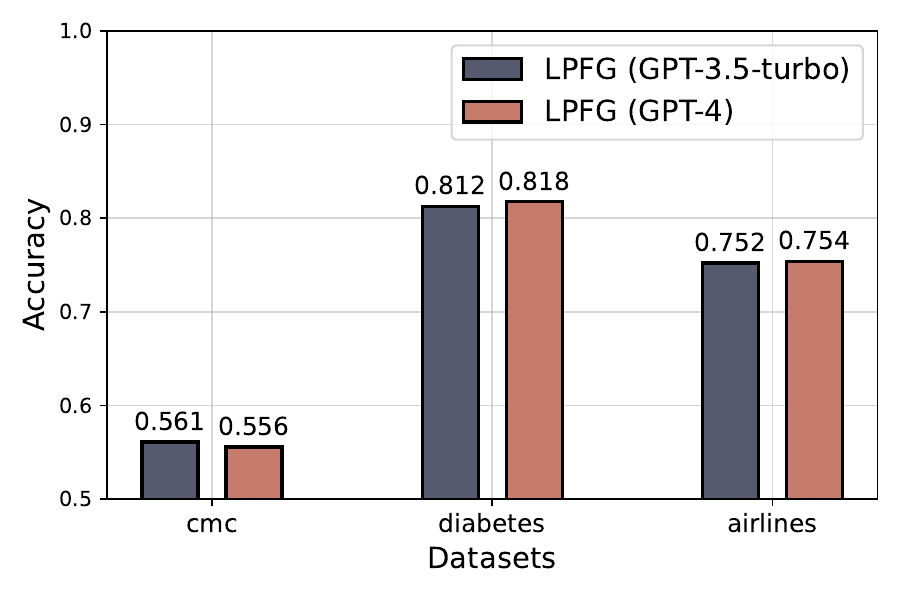}
    \label{exp:llm}
    \end{minipage}
    }
    \centering
    \subfigure[Guidance Impact]{
    \begin{minipage}[ht]{0.472\linewidth}
    \centering
    \includegraphics[width=1.6in]{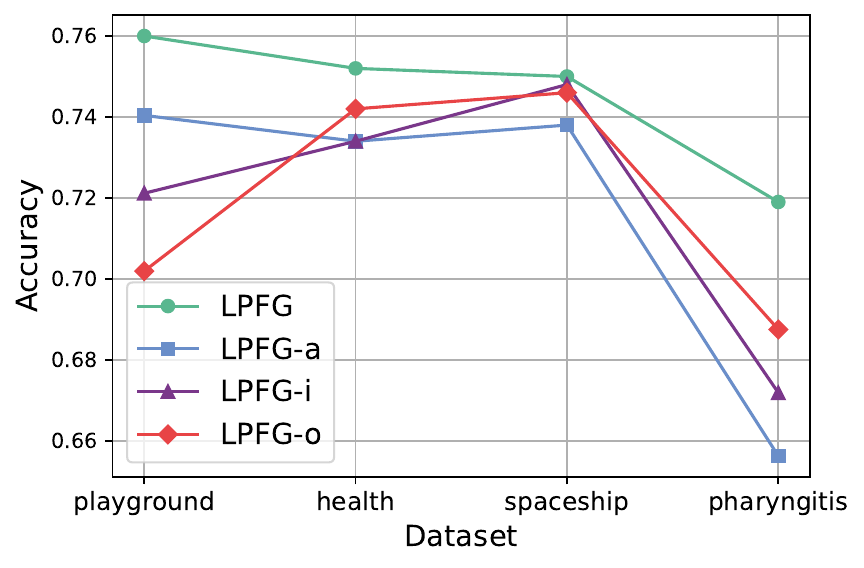}
    \label{exp:optimizer}
    \end{minipage}
    }
    \caption{Ablation study. (a) We study the impact of using different LLMs in LPFG. (b) We investigate the performance of generator guided by different information.}
    \vspace{-0.4cm}
    \label{exp:ab}
\end{figure}

\noindent\textbf{The Impact of Different LLMs} This experiment aims to answer: \textit{Does the choice of LLM model affect the performance of our method?}
We adopt different LLMs to build the system and analyze the influence of the LLM backbone on performance. Figure \ref{exp:llm} presents the results of LPFG with \textit{GPT-3.5-turbo} and \textit{GPT-4o} respectively. The results indicate that the performance of both models remains comparable. Despite substituting with a more advanced LLM, the accuracy exhibits only marginal variation. The reason is two-fold: For the critic agent, evaluating the dataset from semantic and distributional perspectives is a relatively straightforward task, as the general knowledge embedded in the LLM often enables it to provide constructive suggestions. For the generator, its role is merely to follow the critic agent’s guidance and explore potential feature combinations. Consequently, high-level intelligence is not required, making the implementation of LPFG more cost-effective and practical. 

\noindent\textbf{The Impact of Guidance for Generation} This experiment aims to answer: \textit{Is the advice from critic agent better than supervised signals (e.g., accuracy and feature importance) for guiding generation?}
To validate the effectiveness of the critic agent in facilitating feature transformation, we introduce variant models LPFG-a and LPFG-i, which utilize downstream model accuracy and feature importance as guidance, respectively, while LPFG-o operates without any feedback. 
% As shown in Figure \ref{exp:optimizer}, across all four datasets, the performance of LPFG without the optimizer exhibits a noticeable decline. This is because feature transformation is a complex process involving the selection and combination of features and operators. A single generator struggles to thoroughly analyze data patterns and produce reasonable features. The proposed optimizer-generator can be regarded as a cascade framework. We first analyze data patterns, feature interactions, and task objectives, and then make interpretable decisions based on the extracted information. 
As illustrated in Figure \ref{exp:optimizer}, the first key observation is the noticeable performance decline of LPFG-o compared to LPFG. A underlying driver is that a single LLM struggles to balance reasoning and generation, making it difficult to generate meaningful features. Furthermore, models guided by accuracy and feature importance also experience performance degradation, even underperforming compared to models without any guidance on several datasets. This can be attributed to the generator agent’s inability to effectively interpret non-traceable information within a few-iteration feature transformation process, thereby restricting its capacity to guide the generation of new features. In contrast, the actionable advice generated by the critic agent effectively aids in optimizing the feature space without requiring extensive time for feedback interpretation and potential space exploration.

\begin{figure}[h]
    \centering
    \includegraphics[width=0.7\linewidth]{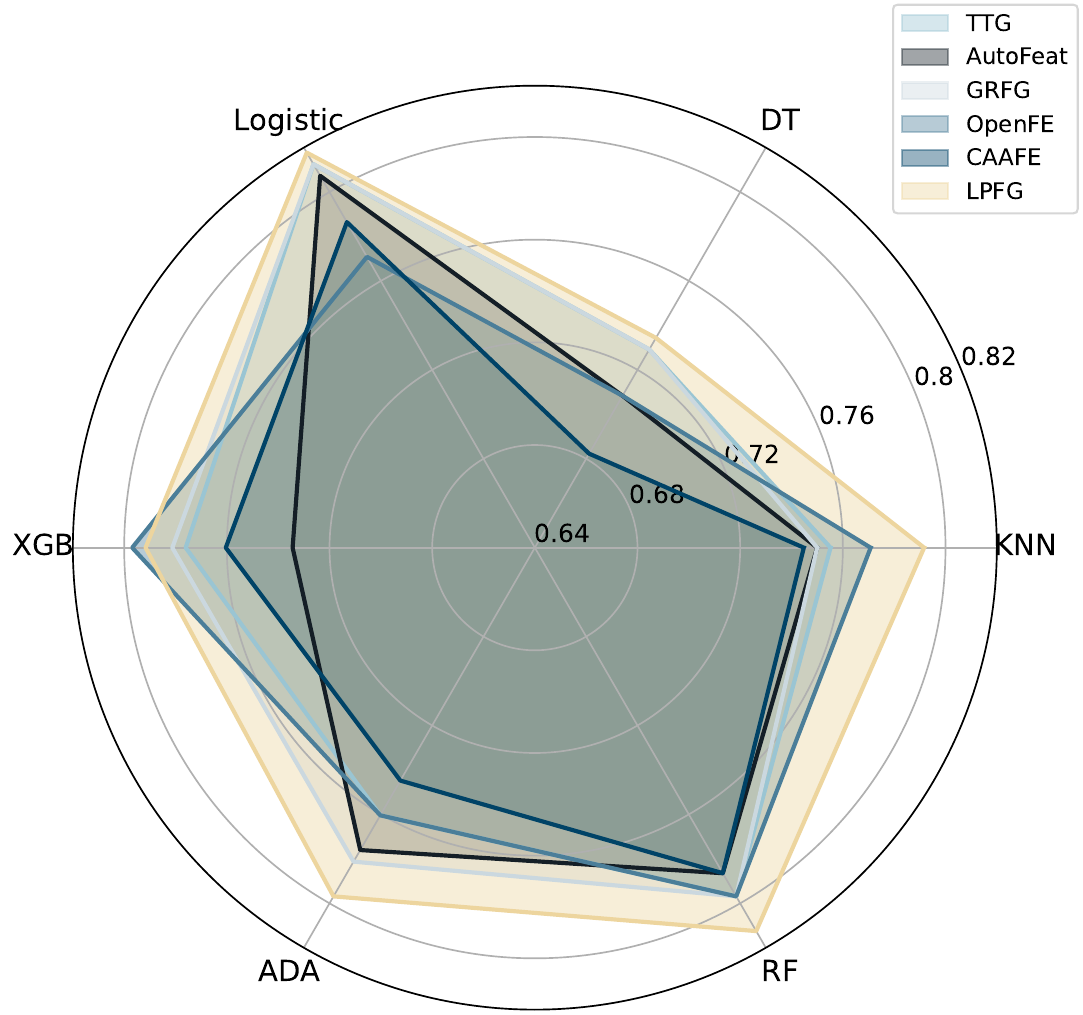}
    \caption{Robustness check. On \textit{diabetes}, we investigate the robustness of the proposed method when different downstream ML models are employed.}
    \label{exp:robust}
    \vspace{-0.4cm}
\end{figure}

\noindent\textbf{Robustness Check} This experiment aims to answer: \textit{Is our method robust when collaborate with different downstream models?}
We employ various downstream ML models, i.e., XGBoost (XGB), Support Vector Machine (SVM), K-Nearest Neighborhood (KNN), Decision Tree (DT), AdaBoost (ADA),  to study the robustness of the proposed method on the dataset \textit{diabetes}. Figure \ref{exp:robust} illustrates that LPFG achieves the best performance, except when the downstream task employs XGB. This can be explained by LPFG is task-agnostic, as it operates in an unsupervised manner.  In contrast, supervised methods are affected by their performance on specific models, leading to greater fluctuations. As an unsupervised plug-in model, LPFG's robustness highlights its practical value in real-world machine learning tasks.

\begin{figure}[h]
    \centering
    \includegraphics[width=0.8\linewidth]{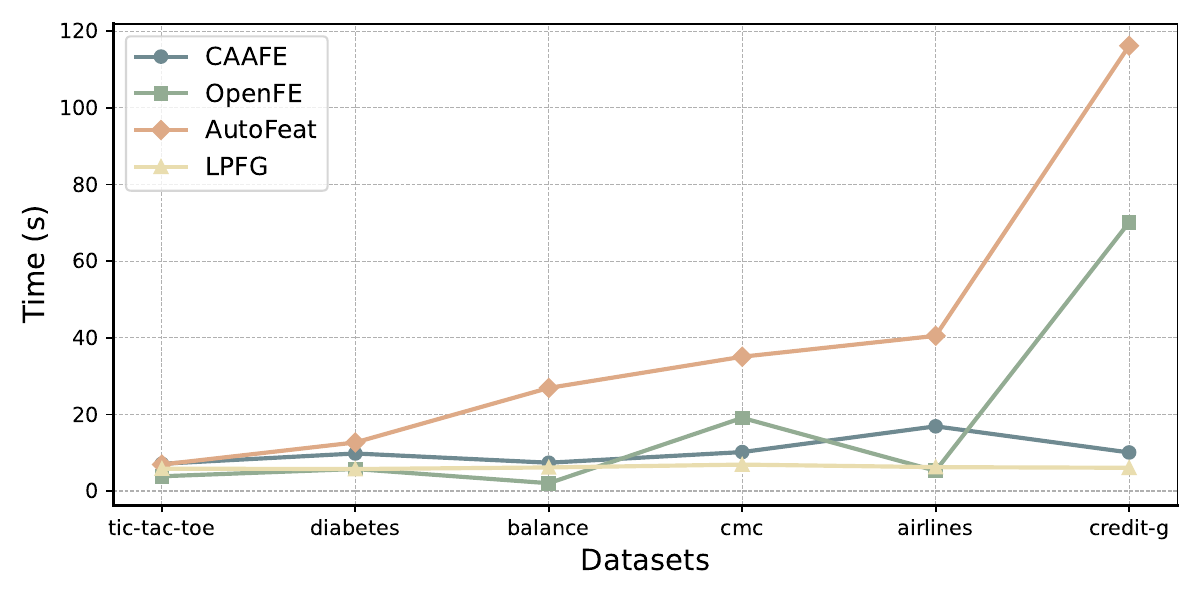}
    \caption{Time complexity study. We compare the costing time (s) for feature tran on 6 datasets.}
    \label{exp:time}
    \vspace{-0.4cm}
\end{figure}

\noindent\textbf{Time Complexity Study} \textit{Is our model efficient in feature transformation?}
We further investigate the time complexity of the proposed method on 6 datasets. Since GRFG and TTG are reinforcement learning-based methods that require time for search within the solution space, we primarily focus on comparing the time complexity of lightweight models. Figure \ref{exp:time} presents the feature transformation time for different models. It can be observed that LPFG demonstrates consistently short and stable generation time. In contrast, the time cost of AutoFeat and OpenFE significantly fluctuates with the size of the dataset. As another LLM-based model, CAAFE requires slightly more time and exhibits minor fluctuations. This can be explained by the fact that LPFG avoids repeatedly computing downstream model accuracy, and the inference time of the LLM through API is fast and consistent.

% \begin{table}[h]
%     \centering
%     \resizebox{\linewidth}{!}{
%     \begin{tabular}{c|cccccc}
%         \toprule
%         & balance & cmc & credit-g & diabetes & tic-tac-toe & airlines \\
%         \midrule
%         CAAFE & 7.3817 & 10.1578 & 10.0577 & 9.7948 & 7.1032 & 16.8760 \\
%         OpenFE & 2.0359 & 19.0776 & 70.0968 & 5.7367 & 3.8435 & 5.2882 \\
%         AutoFeat & 26.8879 & 35.0329 & 116.1725 & 12.6666 & 6.9159 & 40.4818 \\
%         \midrule
%         LPFG & 6.1462 & 6.8919 & 6.0606 & 5.7350 & 5.8252 & 6.2275 \\
%         \bottomrule
%     \end{tabular}}
%     \caption{Time complexity study. We report the costing time (s) of lightweight models on 6 datasets.}
%     \label{exp:time}
% \end{table}

\begin{figure}[h]
    \centering
    \includegraphics[width=\linewidth]{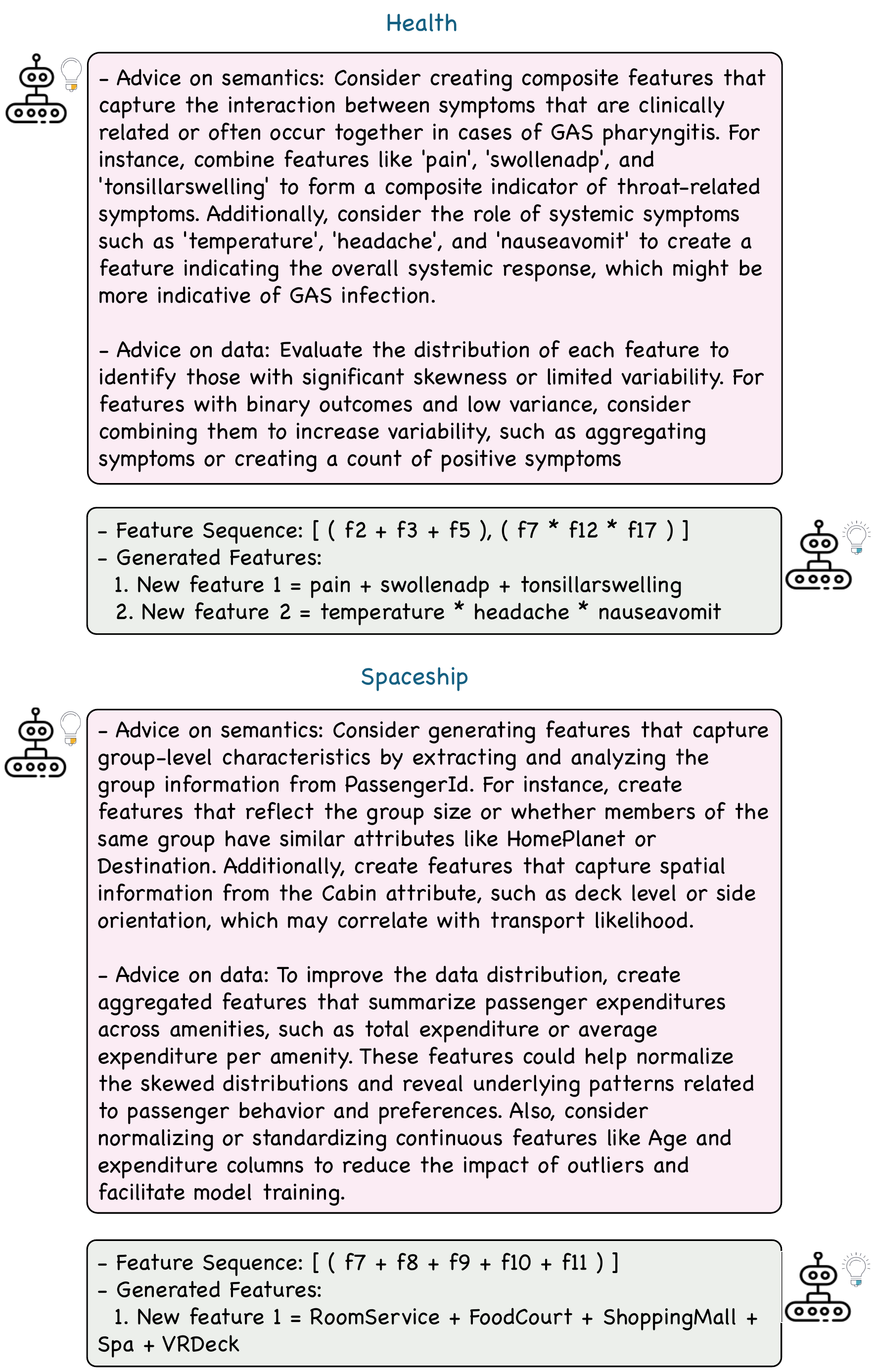}
    \caption{Case study. We provide some examples of generator-critic generation process. The critic agent suggests optimization paths, and the generator agent selects and takes actions on the most promising one.}
    \label{exp:case_1}
    \vspace{-0.4cm}
\end{figure}

\noindent\textbf{Case Study: Duet-Play Generator-Critic Framework for Feature Transformation} This experiment aims to answer: \textit{How can the teaming LLMs collaborate on feature transformation?} We illustrate the communication process between the critic agent and the generator agent in Figure \ref{exp:case_1}. The critic agent provides actionable suggestions for optimizing the feature space, while the generator agent considers these suggestions to make the final decision. Through these examples, we observe two key points: consistency and interpretability. On one hand, the generated features align with certain parts of the advice, indicating that the critic agent provides meaningful guidance in feature transformation. On the other hand, our approach not only enables a concise feature set representation through feature sequences with minimal tokens but also allows for the interpretation of the generated features, enhancing transparency and understanding.

\begin{figure}[h]
    \centering
    \includegraphics[width=0.9\linewidth]{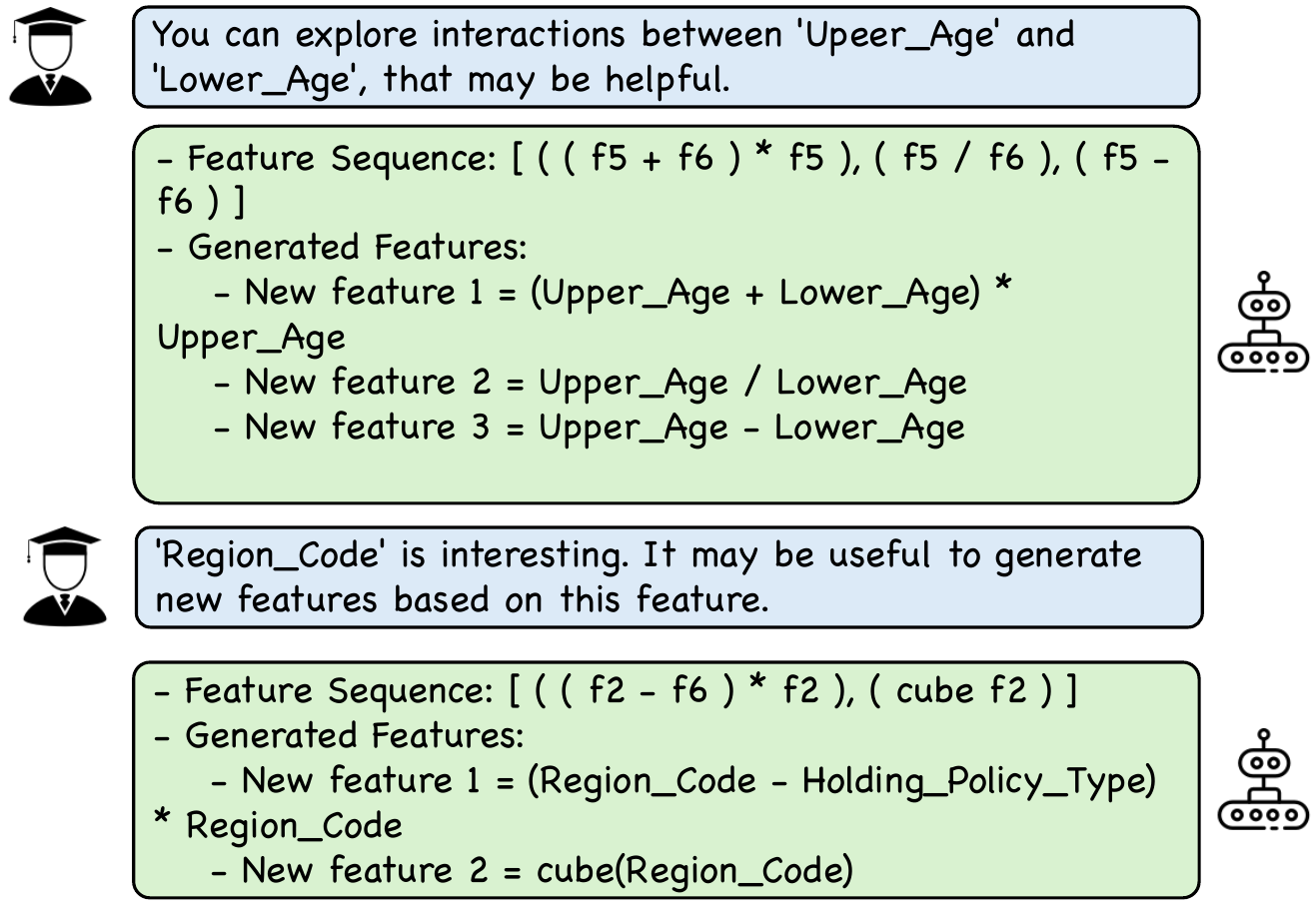}
    \caption{Case study. We provide some examples of conversational feature transformation. LPFG can create new features according to personal requirements.}
    \label{exp:case}
    \vspace{-0.4cm}
\end{figure}

\noindent\textbf{Case Study: Conversational Feature Transformation} This experiment aims to answer: \textit{How can users achieve customized feature transformation in a conversational manner?}
We present several cases of conversational feature transformation in Figure \ref{exp:case} to explore the effectiveness and flexibility of our framework. Instead of adopting an automatic critic agent, we proactively input feature transformation requirements or suggestions, allowing the generator to operate in a more customized manner. This interactive and adaptive approach offers a novel solution for feature transformation, enhancing the practicality and engagement of LPFG.

\section{Related work}

\subsection{Feature Transformation}
% Feature Generation aims to improve the feature space by performing mathematical operations on original features to enhance downstream models. Previous works are mainly two-fold: 1) Discrete decision-making methods: These methods regard feature transformation as a decision-making task in discrete space. Various strategies can be leveraged to improve the trajectory and search for the optimal solution \cite{kanter2015discrete1,khurana2016discrete2,tran2016discrete3,xiao2023traceable,xiao2024evolutionary2}. Traditional methods generate multiple features to expand the feature space and perform feature selection \cite{katz2016explorekit}. Some works leverage evolutionary algorithm to enhance the search process \cite{zhu2022evolutionary1,gong2024evolutionary}. Reinforcement learning is also used to decide feature generation and continuously optimize through the rewards \cite{wang2022GRFG}. 2) Continuous optimizing methods: These methods embed a feature set as a vector in continuous space and perform optimization in the well-constructed embedding space \cite{wang2024MOAT,zhu2022difer}. However, these methods mainly focus on sequential representation of a feature set, ignoring the original tabular information. This paper proposes to view feature transformation from a multimodal perspective and leverage contrastive learning to enable robust search.

Feature transformation is an essential task in Data-Centric AI \cite{wang2025towards,ying2025survey,wang2024knockoff,hu2024reinforcement,gong2025neuro}. It
aims to generate new features to improve the feature space and enhance downstream models \cite{ying2023self,ying2024unsupervised,hu2024reinforcement}. For example, ExploreKit \cite{katz2016explorekit} creates an extensive set of candidate features by integrating information from the original features and Cognito \cite{kanter2015discrete1} investigates diverse feature construction options using a hierarchical and selective approach. Furthermore, Various strategies for discrete decision-making are applied in this task. EAAFE \cite{zhu2022evolutionary1} proposes to leverage evolutionary algorithm to improve feature transformation. GRFG \cite{wang2022GRFG} designs three reinforcement learning agents to collaborate generating new features. Recently, generative feature transformation is proposed and achieves promising performance. MOAT \cite{wang2023reinforcement} formulates feature transformation as a sequential generation task. They embed the feature sets into a continuous space and perform gradient-steered search for the optimal feature set.

\subsection{Task-specific LLM}

With LLM demonstrating competitive performance across a wide range of fields, there is an increasing number of works focusing on the application of LLM for specific tasks \cite{wang2025mixllm,xie2025transformer}. Aug-iModels~\cite{singh2023augmenting} enhances linear models with LLM embeddings and decision trees with LLM-generated features, improving performance and interpretability in NLP tasks and showing promise in neuroscience. The paper~\cite{li2023large} reviews LLM applications in finance, proposes a decision framework for selecting solutions based on data, compute, and performance, and discuss key limitations to guide responsible financial AI use. LLM is also utilized to mine and understand relationships in graph data and are applied to recommendation tasks \cite{wang2024llm}.
There are also methods utilizing LLM for data science. For instance, CAAFE \cite{hollmann2024large} leverages an LLM to iteratively generate additional semantically meaningful features for tabular datasets using the dataset's description. ELLM-FT \cite{gong2024evolutionary} proposes to integrate evolutionary algorithm and LLM to generate new feature sets by few-shot prompting. This paper differs from the previous LLM-based methods in two key aspects: 1) it is under an unsupervised setting which is very challenging for feature transformation; 2) we leverage two specialized LLM agents for dataset diagnosis and feature generation, rather than a single LLM. 3) we extend the method to conversational feature transformation, providing a novel interactive way for this task.

\section{Conclusion}
% We introduce an interpretable language prompting LLM-based unsupervised feature generation model. Our approach implements unsupervised feature generation in two pipelines: 1) we construct an optimizer-generator framework. The optimizer analyzes the feature interactions, semantic relationships of the datasets and provides reasonable suggestions on feature generation. The generator leverages its general knowledge and the advice to create new features and enhance the datasets. 2) By replacing the optimizer with an expert/user, we achieve customized feature generation in a conversational manner. Through extensive experiments, we confirm the effectiveness, efficiency and practical value of LPFG.

We introduce a duet-play generator-critic LLM agents model. Our approach implements unsupervised feature generation in three steps: 1) we employ a critic agent for dataset diagnosis. Leveraging the general knowledge, it provides feature set improvement suggestions from both semantic and distributional perspectives in an unsupervised manner; 2) we build a generator agent to create new features. Based on the advice from the critic agent, the generator makes the final feature set optimization decision and generates new features in a sequential formulation; 3) We iterate the feedback loop between the critic agent and generator agent, continuously refining the feature space. The proposed framework achieves unsupervised dataset diagnosis and improvement. By teaming two specialized LLM agents, we avoid repeated feature combination space exploration and implement robust and efficient feature set optimization in few iterations. Our method is also extended to a novel conversational feature generation formulation. Replacing the critic agent with a human expert, we integrate the expertise into LLM and build a flexible and interactive system for feature generation. Finally, extensive experiments demonstrate the effectiveness, robustness, efficiency, and traceability of our method.

%% The file named.bst is a bibliography style file for BibTeX 0.99c
\bibliographystyle{named}
\bibliography{refer}

\end{document}